\pdfoutput=1

\documentclass[11pt]{article}
\usepackage[]{acl}

\usepackage{times}
\usepackage{latexsym}
\usepackage[T1]{fontenc}
\usepackage[utf8]{inputenc}
\usepackage{microtype}
\usepackage{inconsolata}
\usepackage{todonotes}
\usepackage{bbding}
\usepackage{soul}
\usepackage{xcolor,colortbl}
\usepackage{appendix}
\usepackage{enumitem}
\usepackage{booktabs}
\usepackage{algorithm}
\usepackage{algorithmic}
\usepackage{multirow}
\usepackage{array}
\usepackage{arydshln}

\definecolor{LightGreen}{rgb}{0.8,1,0.89}
\definecolor{LightRed}{rgb}{1.0,0.8,0.7}
\definecolor{LightCyan}{rgb}{0.88,1,1}
\usepackage{amsmath,amsfonts}
\usepackage{graphicx}
\usepackage{lipsum}

\newcommand{\model}{\texttt{PIECE}}

\title{Knowledge Planning in Large Language Models\\for Domain-Aligned Counseling Summarization}

\author{Aseem Srivastava\textsuperscript{$1$}, Smriti Joshi\textsuperscript{$2$},   Tanmoy Chakraborty\textsuperscript{$3$}, Md. Shad Akhtar\textsuperscript{$1$}\\
        \textsuperscript{$1$}IIIT Delhi, \textsuperscript{$2$}Wysa, \textsuperscript{$3$}IIT Delhi\\ 
        \{\texttt{aseems}, \texttt{shad.akhtar}\}\texttt{@iiitd.ac.in}, \texttt{smriti@touchkin.com}, \texttt{tanchak@iitd.ac.in}
    }

\begin{document}
\maketitle

\begin{abstract}
    In mental health counseling, condensing dialogues into concise and relevant summaries ({\em aka} counseling notes) holds pivotal significance. Large Language Models (LLMs) exhibit remarkable capabilities in various generative tasks; however, their adaptation to domain-specific intricacies remains challenging, especially within mental health contexts. Unlike standard LLMs, mental health experts first plan to apply domain knowledge in writing summaries. Our work enhances LLMs' ability by introducing a novel {\em planning engine} to orchestrate structuring knowledge alignment. To achieve high-order planning, we divide knowledge encapsulation into two major phases: (i) holding dialogue structure and (ii) incorporating domain-specific knowledge. We employ a planning engine on Llama-2, resulting in a novel framework, \model. Our proposed system employs knowledge filtering-cum-scaffolding to encapsulate domain knowledge. Additionally, \model\ leverages sheaf convolution learning to enhance its understanding of the dialogue's structural nuances. We compare \model\ with $14$ baseline methods and observe a significant improvement across ROUGE and Bleurt scores. Further, expert evaluation and analyses validate the generation quality to be effective, sometimes even surpassing the gold standard. We further benchmark \model\ with other LLMs and report improvement, including Llama-2 $(+2.72\%)$, Mistral $(+2.04\%)$ and Zephyr $(+1.59\%)$, to justify the generalizability of the planning engine.
\end{abstract}

\begin{figure}[t]
  \centering
  \scalebox{1.01}{\includegraphics[width=\columnwidth]{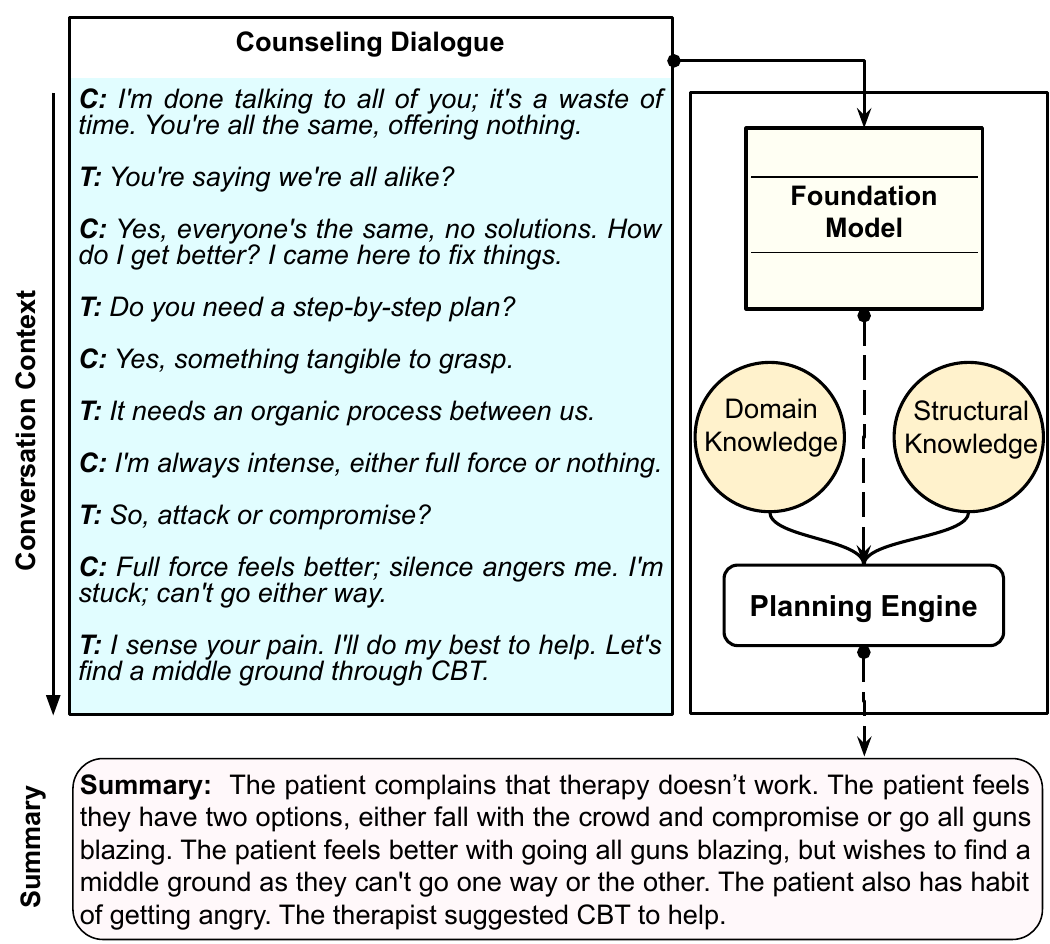}}
  \vspace{-5mm}
  \caption{The proposed pipeline allows LLMs to first plan and then generate. In our approach, prioritizing planning before generation enriches summarization with conversational structure and domain knowledge.}
  \label{fig:page1example}
  \vspace{-5mm}
\end{figure}

\section{Introduction}
Mental health counseling serves as a crucial frontline defense against mental illness. In a typical counseling session, clients articulate their issues while therapists provide support. An essential component of these sessions involves building a strong therapeutic bond and documenting the entire dialogue, commonly known as counseling note, record, or summary\footnote{www.apa.org/gradpsych/2007/01/track}. However, traditional methods of note-taking during sessions can present significant challenges, where therapists are required to divert their attention to note-taking, disrupting counseling interaction and support quality. This distraction not only hinders the therapist's focus but also deteriorates the required therapeutic bond, necessitating the need to automate this process.  

While recent advancements in Large Language Models (LLMs) for mental health are enormous \cite{llmMedicineNature, 10.1001/jamapsychiatry.2023.1253}, it is crucial to incorporate fundamental domain knowledge and understanding of the structural nuances of the dialogue, which current research lacks in ability \cite{cascellaJmedSys}. 
Consequently, we require a knowledge planner to adeptly capture domain intricacies and determine generation priorities. For instance, Figure \ref{fig:page1example} illustrates a sample counseling conversation between a therapist (T) and a client (C). Unlike conventional approaches that directly generate summaries using foundational models, our proposed pipeline introduces a planner. Through this planner, we infuse nuanced dialogue structure and domain-specific knowledge into LLMs.

Earlier efforts in counseling summarization  \citep{10.1145/3534678.3539187} focused on utilizing domain knowledge by integrating patient health questionnaires (PHQ) and counseling components. However, it fell short compared to recent LLMs. The broader scope of such models has been explored for incorporating knowledge into generated text \cite{feng2023trends} and controlling text generation \cite{yu2022survey}. While these efforts are focused on general-purpose downstream tasks, the complexity of counseling dialogue necessitates introducing a more (a) domain-centric and (b) knowledge-structured approach to cater to this problem.  

Our study introduces a novel {\em planning engine}, designed specifically to guide the generation of LLMs. Focused on enhancing counseling summarization, we employ MentalLlama as the foundation to develop our framework -- \model, {\bf p}lanning eng{\bf i}ne for m{\bf e}ntal {\bf c}ounseling not{\bf e} generation. Our model's planning engine exploits knowledge scaffolding and sheaf learner by integrating domain-specific and structural knowledge into LLMs. We evaluate our \model\ against several LLMs, such as Mistral, Zephyr, and Llama. To assess the effectiveness of our approach, we compare \model's performance against $14$ baseline methods. Our quantitative evaluation consists of four automatic summary evaluation metrics -- ROUGE: R-1, R-2, R-L, and Bluert, along with a domain-centric metric, Mental Health Information Capture (MHIC) proposed by \citep{10.1145/3534678.3539187}. We observe a clear improvement of $3.42\%$, 10.11\%, and 6.01\% across R-1, R-2, and R-L metrics, respectively. 
\textcolor{black}{Furthermore, to assess the generalizability of \model, we experiment on ACI-BENCH, a clinical note generation dataset \cite{aci-bench}. We observe that \model\ surpasses the state-of-the-art LLMs with planning-engine on ACI-BENCH as well.}

Additionally, we perform an extensive expert evaluation through an established clinical relevance framework on a set of six dedicated metrics and three task-relevant survey questionnaires. The results affirm the superiority of \model\ against the baselines and demonstrate the adaptability of the planning engine across various LLMs.
Our contributions are summarized below:
\begin{itemize}[leftmargin=*,noitemsep,nolistsep]
    \item We propose \model\ that integrates the planning engine with MentalLlama to address the issue of unreliable generation by LLMs. This engine plans the LLM's generation by filtering dialogue and injecting domain and structural knowledge.
    
    \item We extensively evaluate \model\ against 14 baseline methods. We present \model's significant improvement evaluated across both automatic, human, and expert evaluation metrics. 

    \item We demonstrate the adaptability of the planning engine as it seamlessly integrates with alternative LLMs like Mistral, Zephyr, and Llama. This may also expand the research in planning the generations of diverse LLMs.
\end{itemize}

\noindent \model\ is open sourced for research purpose at \url{https://github.com/LCS2-IIITD/PIECE}

\section{Related Work}
We present our literature review under three major segments to understand and build a planner for LLM-based counseling summarization.

\paragraph{Generative AI in Mental Health Domain.} 
Recent advances in generative research in mental health have led to a diverse range of investigations. For instance, \citet{ashishNature, 10.1145/3442381.3450097} utilized GPT-2 for reinforced feedback generation within peer counseling setups. Subsequent studies employed similar methods with language models (LMs) for empathetic generation \cite{10.1145/3442381.3450097} and facilitated human-AI collaboration. Several studies have concentrated on controlled dialogue generation using LMs, leveraging reinforcement learning \cite{10.1145/3477495.3531912, saha-etal-2022-shoulder, ijcai2023p686, aseemResponseAct, mishra-etal-2023-e, Mishra_Priya_Ekbal_2023, mishra-etal-2023-pal}. Despite this, there remains a scarcity of research aimed at enhancing LLMs in this domain \cite{liu-etal-2023-task}. The concept of counseling summarization was previously explored by \citet{10.1145/3534678.3539187}, where they proposed a filtering mechanism based on annotation and domain knowledge to generate counseling summaries using LMs. As the focus has gradually shifted towards LLMs, recent findings by \citet{yang-etal-2023-towards} have indicated that while LLMs exhibit robust capabilities, they still exhibit significant gaps compared to domain-specific methods.

\paragraph{Knowledge Enhancement in LLMs.} 
There has been a significant effort to incorporate external and domain-specific knowledge into LLMs, as noted in various studies \cite{jiang-etal-2020-know, jiang-etal-2021-know, choi-etal-2023-llms}. Tailoring LLMs to specific domains has notably enhanced their ability to handle downstream tasks. Within the clinical domain, a series of specialized LLMs have emerged \cite{llmClinical}. There exist a few dedicated LLMs specifically trained on mental health corpora, such as the variants of Llama-2 \cite{llama2}, BART \cite{bart}, and T5 \cite{rael_exploring_nodate} as MentalLlama, MentalBART and MentalT5, respectively \cite{mentallama}. While domain-specific LLMs facilitate more contextual relevance, they do not entirely mitigate risks associated with hallucinations or missing context. Planning LLM generations is one of the many solutions here \cite{valmeekam2023on}. Planning methodologies have been explored in various generative tasks, including reasoning \cite{wang2024guiding}, temporal generation \cite{zhang2022generative}, and code generation \cite{zhang2023planning}. However, existing planning systems often lack domain-specific planning, notably within the mental health space.

\paragraph{Text Summarization.}
Knowledge-guided summarization has long been a research focus, and its variations include information-aware techniques \cite{10.1145/3357384.3358020}, perspective-based strategies applied in educational dialogues \cite{jain-etal-2023-summarize} and scientific document summarization \cite{10.1007/978-981-99-8088-8_22, 10.1145/3580305.3599830}, commonsense-driven clinical summarization \cite{10.1145/3583780.3614870}, as well as approaches focusing on topic-awareness \cite{10192022} and attention mechanisms \cite{10261260}. However, many of these methods largely rely on annotated corpora and the LMs to generate coherent text, neglecting deeper exploration into structural comprehension of input data. Limited research efforts, exemplified by \citet{bodnar2022neural}, discuss the use of sheaf for enhanced structural understanding, while \citet{atri-etal-2023-promoting} proposed sheaf to encapsulate structural information for summarization.

\begin{figure*}[ht]
  \centering
  \includegraphics[width=1.0\textwidth]{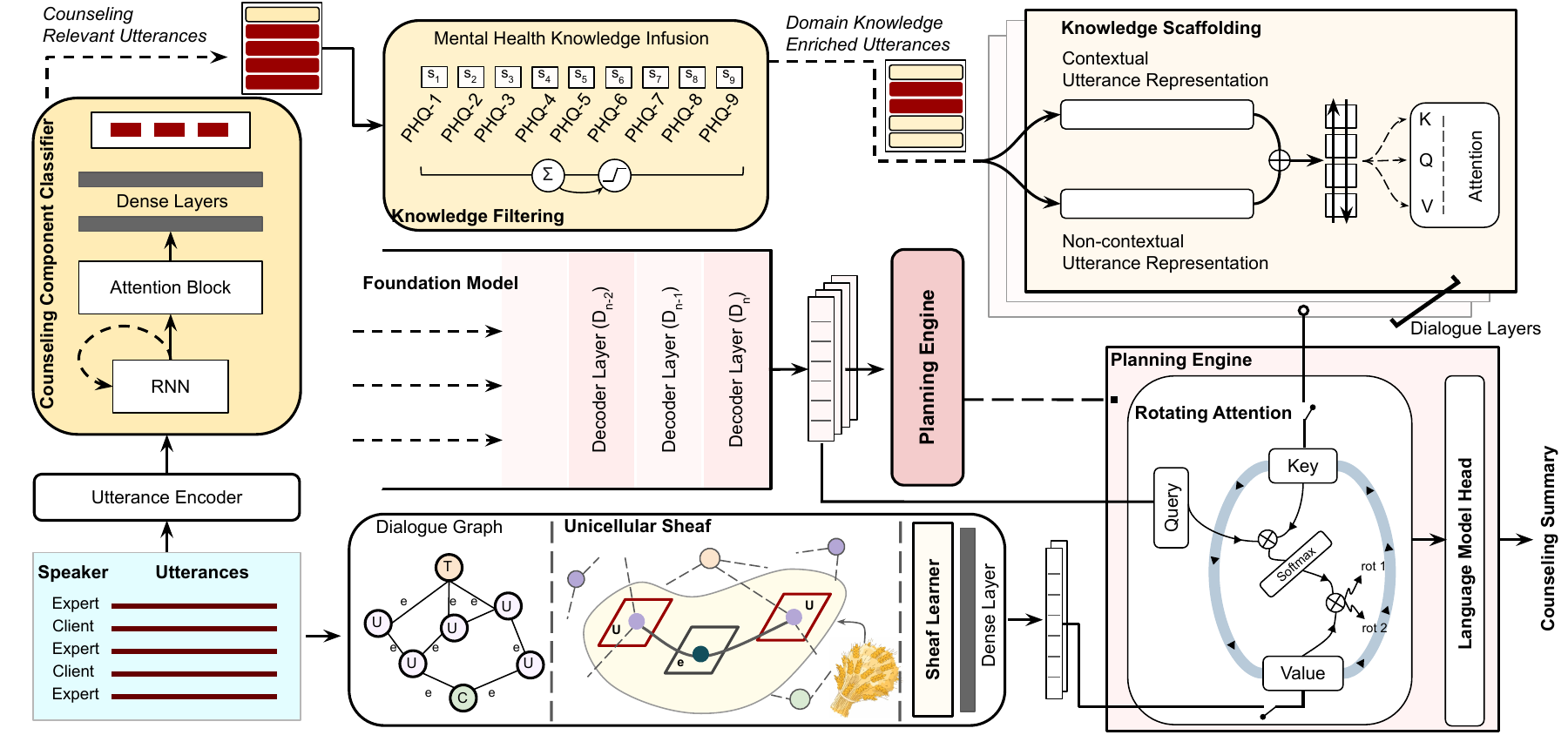}
  \vspace{-5mm}
  \caption{Architecture of \model. We propose a novel {\em planning engine} consisting of two primary sections: (a) integrating knowledge filtering-cum-scaffolding and (b) encapsulating structural understanding of dialogues. The filtration of relevant utterances utilizes counseling component labels within the MEMO dataset to mask filler utterances, followed by knowledge scaffolding. Additionally, sheaf learners are employed for the structural understanding of counseling dialogue. The planning engine operates using a rotating attention mechanism using knowledge from both segments for better LLM generation.  
  }
  \label{fig:model-architecture}
  \vspace{-5mm}
\end{figure*}

\section{Dataset}\label{sec:dataset}
We utilize MEMO, a common counseling summarization dataset with 191 counseling dialogues with $11,543$ utterances \cite{10.1145/3534678.3539187}. Each dyadic counseling dialogue contains a conversation between a therapist ($5,722$ utterances) and a client ($5,814$ utterances), along with an expert annotated counseling summary. In addition to this, MEMO further contains each utterance labeled with a counseling component --  Symptom and History (SH), Patient Discovery (PD), Reflecting (RT), or Discussion Filler (DF) (c.f. Appendix \ref{memodef} for definitions). The dataset contains $2379$ utterances tagged with SH, $5428$ utterances tagged with PD, and $1242$ utterances tagged with RT. These counseling components act as labels of relevance for utterances. 

\section{Proposed Methodology}
Our proposed model, \model, operates on the MEMO dataset and aims to generate knowledge-enriched counseling summaries for a counseling dialogue containing $n$ utterances, $D = {\langle u_1, u_2, u_3, \dots u_n \rangle }$. There are two specific kinds of knowledge that \model\ focuses on -- (a) (mental health) domain knowledge and (b) structural understanding of dialogue. The model achieves this knowledge integration through a novel {\em planning engine} that filters and fuses relevant information to guide the generation of knowledge-enriched summaries by the underlying LLM.  Figure \ref{fig:model-architecture} presents a schematic diagram of \model.

\subsection{Domain Knowledge Encapsulation}
Knowledge encapsulation module involves {\em knowledge filtering} and {\em knowledge scaffolding}. 

\paragraph{Knowledge Filtering.}
Knowledge filtering requires carefully identifying the most relevant utterances for crafting a knowledge-rich summary. This process initiates with classifying each utterance $U_i$ into counseling components (SH, PD, RT, and DF). Among these, SH, PD, and RT are considered essential for generating comprehensive summaries, while DF utterances are regarded as non-essential and are therefore masked by leveraging the dedicated {\em Counseling Component Classifier}. This classifier is purposefully crafted to process the context of utterances, ${\langle u_1, u_2, u_3, \dots u_i \rangle }$, through a GRU unit coupled with a self-attention block. Subsequently, a sequence of two dense layers learns these context-rich utterance representations to predict counseling components and filter non-essential discussion fillers.

The subsequent step involves examining the filtered utterances and further scrutinizing them using a mental health knowledge selection module, MH-Know. Similar to the approach opted by \citet{10.1145/3534678.3539187} with the MEMO dataset, our module leverages the lexicon derived from the Patient Health Questionnaire (PHQ-9) to gauge the degree of mental health-specific content embedded within each utterance \cite{kroenke_phq_9_2001}. The similarity scores ($s_i$) are computed between individual utterances and the PHQ-9 lexicons. Subsequently, a predetermined threshold ($T^\circ = 50\%$, in our study) is applied to ascertain whether an utterance merits retention or should be masked. The resulting filtered dialogue, composed of essential and knowledge-rich utterances, then proceeds to the {\em knowledge scaffolding} stage to ensure optimal preservation of the embedded knowledge.

\paragraph{Knowledge Scaffolding.}
To ensure the cohesion and effective organization of the filtered knowledge segments, \model\ employs a knowledge scaffolding module. We adopt the scaffolding methodology proposed by \citet{scaffold}, where the authors incorporated structural information of scientific papers into citations for citation-intent classification. We modify their knowledge scaffolding for dialogue settings by employing a mix of contextual and non-contextual representations for each relevant utterance, as shown in Equation \ref{scaffoldRep}.
\begin{equation}\label{scaffoldRep}
    R_{k}  = Attention[ \overleftrightarrow{\chi} (C(u_{i}) \oplus C^\circ(u_{i}) )]
\end{equation}
Here, the contextual embeddings $C(x)$ are captured through BERT \cite{devlin2019bert}, while the non-contextual embeddings $C^\circ (x)$ are extracted using Glove \cite{pennington-etal-2014-glove}. These representations are then carefully concatenated and fed into a bidirectional LSTM layer $\overleftrightarrow{\chi}$, ensuring the underlying semantic context. The resulting context-rich representations, enriched with domain-specific knowledge, subsequently serve as key, query, and value vectors within a self-attention block. This orchestrated interplay emerges in generating filtered scaffolded representations $R_{k}$ enriched with the essence of extracted knowledge. Before understanding the role of these representations within the planning engine, we first understand the encapsulation of structural knowledge, another crucial component of \model's knowledge-enriched summary generation pipeline.

\subsection{Structural Knowledge Encapsulation}
\label{method:structure}
To ensure that the generated summaries accurately reflect the overall structure and flow of the original dialogue, \model\ incorporates a dedicated structural knowledge module. Sheaf has been long studied for its ability to hold structural knowledge in graphs via sheaf diffusion \cite{cellularSheaf, curry2014sheaves, bodnar2022neural}. With a similar objective, we first construct a dialogue graph $G = (V, E)$ that captures the intricate network of relationships between the utterances within the dialogue $D$. Each utterance $u_i$ is represented as a vertex $v \in V$ within the graph, while edges $e \in E$ are directed toward the progression of dialogue. In the adjacency matrix $A_d$ of the graph $G$ for each dialogue instance $d \in D$  with $o$ utterances, we place each node's feature $f$ with its encoded utterance representations $f \leftarrow bert(u_i)$. As a result, we construct $G$ to preserve the inherent dialogue structure while learning the graph representations via {\em sheaf learner}.  

\paragraph{Sheaf Learner.} Graphs inherit a notion of the neighborhood but not distance or direction. Sheaf theory provides a mathematical framework for imbuing graphs with geometric structure by associating vector spaces, known as stalks, with each node and edge, along with defining restriction maps (which are essentially linear transformations) between the stalks of incident nodes and edges. These restriction maps capture the local-to-global context, ensuring consistency of information across the graph. The resulting structure, comprising the graph along with the decorated stalks and restriction maps, is termed a cellular sheaf \cite{cellularSheaf}. To facilitate the understanding of complex graph relationships, we construct a sheaf, which is a collection of interconnected cellular sheaves. Drawing inspiration from \citet{bodnar2022neural}, we employ a sheaf convolution network ($SCN$) (see Equation \ref{scn}) to learn these sheaves.
\begin{equation}\label{scn}
    R_{scn} = \eta((I - \Delta)(I \otimes W_1)A^{(o \times f)}W_2)
\end{equation}
Here, A is the input to the network, whereas $W_1$ and $W_2$ are learnable weights. $\Delta$ acts as a sheaf laplacian, $I$ is an identity matrix, and $\otimes$ represents direct matrix multiplication. We apply an activation function $\eta$ on top as a $relu$ function. For the resultant representation $R_{scn}$, we employ a dense layer to learn the graph geometry over topological space,
inheriting better structural knowledge. The resultant structure-rich representations $R_s$ act as an input to the planning engine.

\subsection{Planning Engine}
The planning engine is the core module that integrates the structural and domain-specific knowledge using a rotating attention mechanism. Acknowledging the equivalent significance of both structural and domain-specific insights, the rotating (cyclic) attention mechanism aims to retrieve information evenly from each segment. Here, domain-rich scaffolded representations  $R_k$ and structure-rich sheaf representations $R_s$ act as both key ($K$) and value ($V$) once per cycle, yielding two rich dialogue representations. The query, the foundation model's hidden representations, acts as query ($Q$) for both cycles, as shown in Equation \ref{eq:plan}. 
\begin{equation} \label{eq:plan}
  R_{rep} = sm(\frac{QK^T}{\sqrt{d_k}})V \oplus sm(\frac{QV^T}{\sqrt{d_k}})K
\end{equation}
The language model head (LM-Head) operates on top of the fused representations of both dialogue representations to generate a counseling summary. 

\section{Experiments and Results}
Here, we discuss the selection of baseline methods followed by the performance comparison with baselines, the ablation of \model, and analyses.

%There is a crunch on the dataset's availability for the counseling summarization task. Our experiments are performed on a common counseling summarization dataset named MEMO. We benchmarked the proposed model across $14$ potential baselines. We discuss these details below.

\subsection{Baselines}
We choose the following systems as our baselines --
(i) \textbf{Segmented Modeling (SM)} leverages dialogue-acts on BiLSTM \cite{goo_abstractive_2018}.
(ii) \textbf{BART} is a standard transformer with a BERT encoder and GPT decoder \cite{bart}.
(iii) \textbf{Pegasus} is pretrained with an objective of gap-sentence-generation \cite{zhang_pegasus_2020}. 
(iv) \textbf{T5}  uses a shared framework on transformer to pretrain on huge C4 corpus \cite{rael_exploring_nodate}. 
(v) \textbf{Pretrained Language Model (PLM)} uses DialoGPT to segment topics and generate summaries with BART \cite{feng-etal-2021-language}. 
(vi) \textbf{RankAE} uses an encoder to filter dialogue utterances into segments and generates summaries using denoising auto-encoder \cite{Zou_Lin_Zhao_Kang_Jiang_Sun_Zhang_Huang_Liu_2021}. 
(vii) \textbf{Summ$^N$} is a language model pretrained for dialogue summarization \cite{summn}.
(viii) \textbf{DialogLM} is pretrained for long conversational summarization task \cite{dialoglm}.
(ix) \textbf{ConSum} is the reported state-of-the-art for counseling summarization task on the MEMO dataset, marking as the most relevant baseline method to compare with \cite{10.1145/3534678.3539187}.
(x) \textbf{Flan-T5} is a T5-based LLM, instruction-tuned on a mixture of tasks \cite{flant5}.
(xi) \textbf{Mistral-7B} is an LLM instruction tuned with a sliding window attention mechanism for efficient and longer context  \cite{mistral}.
(xii) \textbf{Zephyr-7B} is a fine-tuned Mistral LLM to surpass the large chat models \cite{zephyr}.
(xiii) \textbf{Llama-2-7B} is an optimized auto-regressive LLM currently state-of-the-art for many language generation tasks \cite{llama2}.
(xiv) \textbf{MentalLlama} is a standard Llama-2 model pretrained on a huge Reddit-based mental health corpus \cite{mentallama}. Our work is proposed on top of MetalLlama because of its domain-specific knowledge and performance.

To evaluate the performance, we employ widely recognized metrics: ROUGE and Bleurt Score.

\begin{table}[t]
\centering
% \vspace{-3mm}

\resizebox{\columnwidth}{!}{
\begin{tabular}{lrrrr}\toprule
% \multirow{2}{*}{Models} &\multicolumn{5}{c}{Result on \data } \\
% \cmidrule{2-6}
{\bf Model} &{\bf R-1} &{\bf R-2} &{\bf R-L} &{\bf BS} \\
\cmidrule{1-5}
SM \cite{goo_abstractive_2018} & 20.46 & 3.80 & 18.87  & -0.9454 \\
BART \cite{bart} &34.92 &12.66 &18.83 &-0.7118\\
Pegasus \cite{zhang_pegasus_2020} & 29.71 & 7.77 & 27.57 & -0.6130\\
T5 \cite{rael_exploring_nodate}&31.44 &5.63 & 27.38 & -0.5655\\
PLM \cite{feng-etal-2021-language}   &34.24 & 11.19 &33.35 & -0.8678\\
RankAE \citet{Zou_Lin_Zhao_Kang_Jiang_Sun_Zhang_Huang_Liu_2021} & 25.57 & 3.43 & 24.16 & -1.0630\\
SUMM$^N$ \cite{summn} &34.06 &11.32 &20.99 &-0.6088\\
DialogLM \cite{dialoglm} &28.14 &9.21 &17.57 &-0.7377\\
ConSum \cite{10.1145/3534678.3539187} &45.36 &15.71 & 24.75 &0.3407\\
% \cdashline{1-5}
Flan-T5 \cite{flant5} &41.30 &16.00 &29.05 &0.2281\\
Mistral-7B \cite{flant5} &46.74 & 14.98 &32.48 &0.3056\\
Zephyr-7B \cite{zephyr}&38.90 &9.90 &26.35 &0.0094\\
Llama-7B \cite{llama2}&47.22 &16.74 &33.16 &{\bf 0.4106}\\
MentalLlama \cite{mentallama}&47.92 &16.90 &35.81 &0.3953\\
\cmidrule{1-5}
\rowcolor{blue!14} {\bf \model}\ ($fm:$ MentalLlama) &\textbf{49.62} &\textbf{18.61} & \textbf{38.10} &0.4102\\
% \cmidrule{1-5}
$\Delta_{\model-BEST}(\%)$ & \textcolor{blue}{$\uparrow 3.42$} & \textcolor{blue}{$\uparrow 10.11$} &\textcolor{blue}{$\uparrow 6.01 $} &\textcolor{red}{$\downarrow 0.09$}\\
% \cmidrule{1-5}
% % $\qquad$ -- {\em StructKnow}  -- {\em DomainKnow}   & - & - & - & -\\
% $\quad${\bf Ablations}                        & & & & \\
% $\qquad$ \model\ -- {\em StructKnow}                        & 48.25 & 16.22 & 37.40 & 0.4085\\
% $\qquad$ \model\ -- {\em DomainKnow}                        & 48.36 & 15.78 & 36.96 & 0.3994\\
% \cdashline{1-5}
% $\quad$\model\ ($fm=\alpha$)                             & & & & \\
% $\qquad$ $\alpha:$ Mistral                        & 46.81 & 13.28 & 34.52 & 0.3080 \\
% $\qquad$ $\alpha:$ Zephyr                         & 41.63 & 11.36 & 27.94 & 0.0106\\
% $\qquad$ $\alpha:$ Llama                          & 48.25 & 14.02 & 35.88 & 0.4098\\

\bottomrule
\end{tabular}
}

\caption{Results obtained on the MEMO counseling summarization dataset. We report  Rouge-1 (R-1), Rouge-2 (R-2), Rouge-L (R-L), and Bleurt Score (BS).}
\label{tab:results}
 \vspace{-4mm}
\end{table}

\subsection{Performance Comparison}
Table \ref{tab:results} shows the performance of the baseline models, revealing the clear superiority of \model\ across three out of four metrics. Notably, MentalLlama turns out to be the best-performing baseline. \model\ excels in capturing both semantic and syntactic structures, as evidenced by the improvements of best-performing LLMs like Llama variants. Specifically, \model\ demonstrates improvements of $+3.42\%$, $+10.11\%$, and $+6.01\%$ in R-1, R-2, and R-L points, respectively. On the other hand, Llama yields the best scores for the Bluert metric; however, \model\ closely matches its performance, exhibiting a marginal drop of only $-0.0004$ Bluert points. However, compared with ConSum, the benchmarked state-of-the-art model for counseling summarization,  our findings reveal a significant boost in performance metrics. Notably, \model\ shows improvements of $+9.39\%$, $+18.45\%$, $+53.93\%$ and $+20.39\%$ in R-1, R-2, R-L and Bleurt points, respectively. 

\paragraph{Generalizability.} \textcolor{black}{MEMO is the only publicly available dataset that fits into the definition of counseling summarization, to the best of our knowledge. However, we further add experiments on another clinical note generation dataset, ACI-BENCH \cite{aci-bench}, as shown in Appendix (\S Table \ref{performanceComparison})}.

\subsection{Ablation Study}
We perform an ablation study to assess the performance of underlying components contributing to \model. By systematically deconstructing the planning engine and analyzing various elements of our model architecture, we present our findings in Table \ref{tab:ablation}. The impact of the planning engine on the acting foundation models ($fm$) assesses the generalizability of the planning engine. Evidently, the role of the addition of knowledge is highlighted as we observe a clear decline in the performance of \model\ across all four metrics (c.f. Appendix \ref{strVSdom} for comparative study). We observe the same trend for both structural and domain knowledge, with the performance declining by a significant margin of $-2.83$ (R-2) points. At the same time, we perform a generalizability check for the planning engine to be inducted on top of multiple LLMs such as Mistral, Zephyr, and Llama, finding that each LLM, in general, benefits from the planning engine.

\begin{table}[t]
\centering
% \vspace{-3mm}

\resizebox{\columnwidth}{!}{
\begin{tabular}{lcccc}\toprule
% \multirow{2}{*}{Models} &\multicolumn{5}{c}{Result on \data } \\
% \cmidrule{2-6}
{\bf Ablations} &{\bf R-1} &{\bf R-2} &{\bf R-L} &{\bf BS} \\
\cmidrule{1-5}
\model\ & & & & \\
$\quad$ -- {\em StructKnow} & 48.25 (\textcolor{red}{$\downarrow 1.37$}) & 16.22 (\textcolor{red}{$\downarrow 2.39$}) & 37.40 (\textcolor{red}{$\downarrow 0.70$}) & 0.4085 (\textcolor{red}{$\downarrow 0.0017$})\\
$\quad$ -- {\em DomainKnow} & 48.36 (\textcolor{red}{$\downarrow 1.26$})& 15.78 (\textcolor{red}{$\downarrow 2.83$})& 36.96 (\textcolor{red}{$\downarrow 1.14$})& 0.3994 (\textcolor{red}{$\downarrow 0.0108$})\\
\cdashline{1-5}
\model\ ($fm=\alpha$)                    & & & & \\
$\quad$ $\alpha:$ +Mistral & 46.81 (\textcolor{blue}{$\uparrow 0.07$}) & 13.28 (\textcolor{red}{$\downarrow 1.70$}) & 34.52 (\textcolor{blue}{$\uparrow 2.04$}) & 0.3080 (\textcolor{blue}{$\uparrow 0.0024$}) \\
$\quad$ $\alpha:$ +Zephyr & 41.63 (\textcolor{blue}{$\uparrow 2.73$})& 11.36 (\textcolor{blue}{$\uparrow 1.46$})& 27.94 (\textcolor{blue}{$\uparrow 1.59$})& 0.0106 (\textcolor{blue}{$\uparrow 0.0012$})\\
$\quad$ $\alpha:$ +Llama & 48.25 (\textcolor{blue}{$\uparrow 1.03$})& 14.02 (\textcolor{red}{$\downarrow 2.72$})& 35.88 (\textcolor{blue}{$\uparrow 2.72$})& 0.4098 (\textcolor{red}{$\downarrow 0.0008$})\\

\bottomrule
\end{tabular}
}

\caption{Ablation study of the proposed model, \model\, and generalizability of the planning engine on top of notable LLMs as foundation model ($\alpha$), including Mistral, Zephyr, and Llama, illustrating a clear \textcolor{blue}{improvement ($\uparrow$)} in LLM generations by integrating planning engine.}
\label{tab:ablation}
\vspace{-4mm}
\end{table}

\subsection{Qualitative Analysis}
We assess \model's text generation capabilities, apart from quantitative metrics, via detailed comparative analysis between \model\ and the two most relevant models: Llama and MentalLlama. Given that \model\ builds upon the MentalLlama, we carefully compare its performance against these two models and the gold standard. Table \ref{tab:analysis} presents an illustrative comparison. Our scrutiny centers on discerning the understanding of core counseling components, including discussions on {\em symptoms and reasons}, {\em reflecting client's story}, and {\em critical routine details} of the dialogue along with the {\em additional domain understanding}. 

\begin{table*}
\centering
% \fontsize{6}{7}\selectfont
\resizebox{\textwidth}{!}{%
\begin{tabular}{l|p{58em}} \toprule
\textbf{Model} & {\centering \textbf{Summaries}} \\\cmidrule{1-2}
Gold & {\em Patient visited the therapist for feeling bad. Therapist knew that patient had such feelings in the past, but this time patient found it harder. Two months ago patient visited GP after that GP referred to therapist. Earlier patient had a breakup and felt lonely. Therapist asks how things have been since the patient went to GP. Patient feels nothing changed and unable to accept failure. Patient's mom was not happy with the patient's behavior. Therapist suggested CBT. Patients accept that the first year was tough. Therapist asks that it doesn't sound like you want to go back. Patient wants to but just doesn't feel like meeting the man. Patient used to sit in a friend's restaurant for money and stuff in the meantime. Patient used to like it because the patient used to speak to people. Therapist asks to scale the thoughts out of 100. Patient says 80. Patient feels empty when asked about negative thoughts. Patient doesn't want her mom to be too worried.}\\ 
\midrule

Llama & \textit{Patient adrift after a \textbf{\color{green!60!black}brutal breakup} finds solace in a therapist's office. The weight of past struggles, a university hiatus, and crushing expectations press down, leaving her numb and weary. \textbf{\color{green!60!black}Work, once at a restaurant, becomes a reminder of self-doubt}. But as secrets unfurl – \textbf{\color{red}anxieties, fears, and the facade she wears for her mom} – a glimmer of hope flickers. Maybe, just maybe, this conversation, this courageous first step, can illuminate a path back to wholeness.} \\
\midrule

MentalLlama & {\em Patient, \textbf{\color{red}battling depression and loneliness}, seeks therapeutic guidance. Past \textbf{\color{green!60!black}heartbreak} and \underline{struggles weigh heavily}, \textbf{\color{red}leaving her drained and empty}. \textbf{\color{blue}Work offers a temporary escape}, but anxieties about "\underline{not being good enough}" resurface. The therapist recognizes the patient's inner strength and the \textbf{\color{red}burden of hiding pain from her mother}. Through the conversation, a flicker of hope emerges, hinting at the \underline{possibility of a different path forward} rooted in understanding and compassion.}\\ 
\midrule

\bf \model & {\em The patient visited the therapist feeling \underline{unable to recover from recent emotional difficulties}, notably following a \textbf{\color{green!60!black}breakup with a long-term boyfriend}. \underline{Expressing disappointment in seeking therapy} and \textbf{\color{blue}reluctance to admit feelings of failure after being referred by the GP}, the \textbf{\color{red}therapist highlighted the normalcy of anxiety}. \textbf{\color{blue}Concerns about returning to university after a break were discussed}, \underline{reflecting the patient's desire for success} amidst \textbf{\color{red}overwhelming pressure and self-doubt}. Additionally, feelings of inadequacy arose during a recent \textbf{\color{green!60!black}work shift at a friend's restaurant, causing the patient to feel good but confused in talking to people}. The therapist further explored the \textbf{\color{green!60!black}patient's sense of guilt for not meeting expectations, especially with mom}, \underline{revealing a deep emotional impact}. The \textbf{\color{red}patient admitted to feeling empty and drained}, \textbf{\color{blue}rating their emotional distress at 80 out of 100 and describing physical sensations resembling nervousness}. The therapist understood the \textbf{\color{green!60!black}patient's struggle to appear unaffected for their mother's sake}, \underline{despite internal turmoil, ultimately feeling worse for doing so}.}\\

\bottomrule
\end{tabular}
}

\caption{A comparative analysis of three most relevant models -- Llama, MentalLlama, and \model. The key emphasis to be analyzed here is understanding the core counseling components, including discussions on \textit{{\color{red}symptoms and reasons}, {\color{green!60!black}reflecting client's story}}, and \textit{{\color{blue}critical routine details}} along with the \textit{\underline{additional intricate domain understanding}}. While Llama generates general-purpose summaries, MentalLlama, being pretrained on mental health data, captures nuanced knowledge beyond Llama's scope. In contrast, \model\ is able to capture in-depth domain knowledge surpassing other models in touching upon counseling nuances. Despite the grammatical proficiency of baselines, \model\ stands out for its structural understanding, focusing on intricate details, in some cases, better than gold.}

\label{tab:analysis}
\vspace{-5mm}
\end{table*}

\begin{table}[!b]
% \vspace{-3.5mm}
\centering

\resizebox{\columnwidth}{!}{
\begin{tabular}{l|cccc}\toprule
Model & Relevance & Consistency & Fluency & Coherence \\
\cmidrule{1-5}
Llama  & 3.12 & 3.22 & {\bf 3.76} & 3.68 \\
MentalLlama  & 3.57 & 3.31 & 3.75 & 3.71 \\
\cdashline{1-5}
\textbf{\model} & {\bf 3.73} & {\bf 3.39} & 3.72 & {\bf 3.75}\\
\bottomrule
\end{tabular}
}
\caption{Human evaluation on the summaries generated from \model\ model. The average interrater's agreement score $(\kappa)$ for \model\ is 0.82.}
\label{tab:humaneval}

\end{table}

In Table \ref{tab:analysis}, the first row showcases the gold standard summary. Evidently, Llama excels in text coherence but falls short in producing domain-specific insights and barely touches upon the critical domain information, for example, {\em client's referral from GP, self-emotional-assessment rating, and declaration of anxiety}. In contrast, MentalLlama demonstrates better mental health-specific nuances, highlighting counseling components such as {\em battling depression and loneliness, the burden of hiding pain, and past heartbreaks} and integrating intricate domain information such as {\em struggles weigh heavy and the possibility of a different path forward}. Apparently, MentalLlama's performance lacks a grasp of the structural intricacies of dialogue. On the other hand, the summaries by \model\ are descriptive, emphasizing the complete dialogue structure and crucial counseling components inherited from the conversation such as {\em therapist highlighting normalcy, self-assessment, reluctance to admit feelings}, instances which MentalLlama and Llama skipped. Additionally, \model\ excels in enhancing domain understanding by incorporating spans from the LLM’s vocabulary rather than the conversation’s vocabulary.

\paragraph{Error Analysis:} \textcolor{black}{We extend the analyses for cases where our model falls short in capturing the intended details (c.f. Appendix \ref{errorSection}). We discuss two important cases: a) \model\ tends to include exaggerated information in shorter dialogues, leading to extra details, and b) \model\ summaries are usually longer than gold summaries and, on average, struggle to capture patient’s behavior.}

\paragraph{Human Evaluation}
We present human evaluation on four standard linguistic parameters, namely \textit{relevance} (selection of relevant content), \textit{consistency} (factual alignment between the summary and the source), \textit{fluency} (linguistic quality of each sentence), and \textit{coherence} (structure and organization of summary) as shown in Table \ref{tab:humaneval}. We employed 12 linguistics experts to rate each parameter on the Likert scale of $1$ to $5$. Out of twelve, seven were female, whereas five were male, all of them aged between 23 - 35.
As shown in Table \ref{tab:humaneval}, \model\ surpasses across three out of four parameters. Notably, it achieved a score of $3.73$ for relevance, indicating that \model's summaries capture the core knowledge. Additionally, \model\ scored $3.39$ and $3.75$ for consistency and coherence, respectively, showing the logical flow of the original conversation. Finally, \model\ competes with $Llama$ on {\em fluency} metric. The superior performance underscores the model's linguistic quality and structural coherence. The average Cohen's kappa score ($\kappa$) for \model\ is $0.82$, which falls under the substantial category.

\begin{figure}[t]
  \centering
  \scalebox{1.01}{\includegraphics[width=\columnwidth]{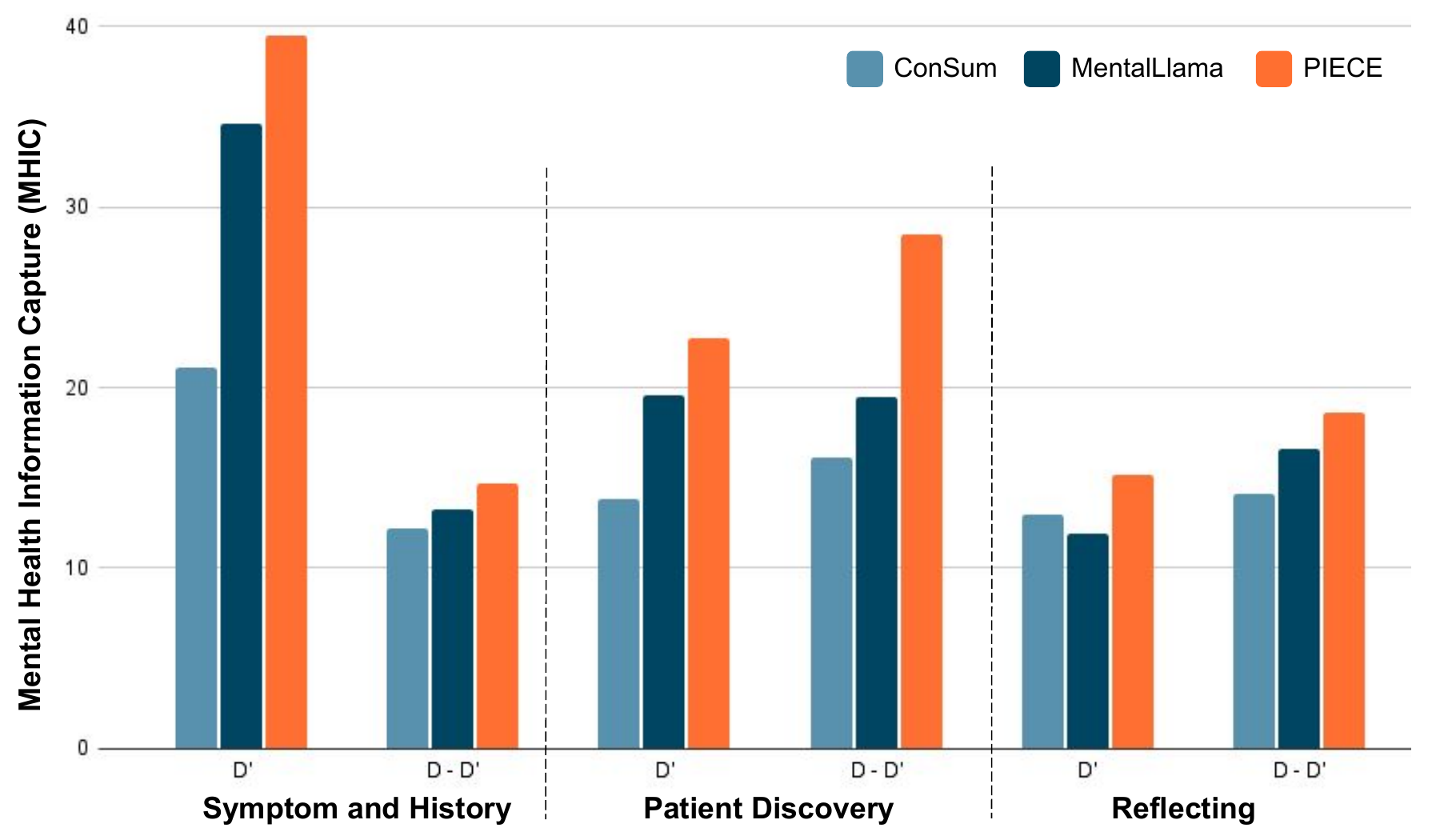}}
  \caption{Domain-centric evaluation using Mental Health Info Capture (MHIC) metric. The proposed model, \model, distinctly excels in capturing domain knowledge compared to the two most relevant models.}
  \label{fig:mhic}
  \vspace{-5mm}
\end{figure}

\subsection{Mental Health Information Capture} 
The Mental Health Information Capture (MHIC) computes the intersection of utterances predicted by the counseling components classifier and the generated summary using the ROUGE-1 score to provide a qualitative evaluation of these summaries. 
Figure \ref{fig:mhic} presents a comparative metric study of the three top-performing models -- ConSum, MentalLlama, and  \model. 
Evidently, the earlier state-of-the-art benchmarked model, ConSum, on the MEMO dataset appears to be easily surpassed by MentalLlama without any additional effort. 
However, after integrating the {\em planning engine} to MentalLlama, i.e., \model, performance surpassed both variants, highlighting the efficacy of \model.

\subsection{Mental Health Expert Validation} \label{expVal}
Our model, \model\, acts as an assistance for complex counseling processes, as experts can utilize the summaries by \model\ as counseling notes, reducing both time and cost. To validate the generation's effectiveness, we collaborated with a clinical psychologist (having 10+ years of clinical experience) who assessed the summaries on two major fronts: clinical acceptability and LLM relevance.

\paragraph{Clinical Acceptability.} Experts exploit a clinical acceptability framework \cite{Sekhon2017} that evaluates parameters such as {\em affective attitude, burden, ethicality, coherence, opportunity costs}, and {\em perceived effectiveness} (c.f. Appendix \ref{expDef} for definitions). The expert rated summaries on a scale from 0 to 2, with higher scores indicating better acceptability (c.f. Table \ref{tab:experteval}). The expert evaluation shows an average rating of $1.20$ out of $2.00$, falling well within the standard acceptability range of $0.70$ to $1.40$. As a result, the expert concluded that the generated summaries demonstrate suitability for therapists, sometimes surpassing the gold standard. 

\paragraph{LLM Relevance.} To assess common LLM flaws, quality of generation, and relevance to experts, we ask experts to address the following aspects:

\begin{itemize}[leftmargin=*,noitemsep,nolistsep]
    \item Did the expert observe our model hallucinate? 
    \item How does the model's generation compare to that of the most competitive model (MentalLlama)?
    \item Is the generated summary relevant to the expert? 
\end{itemize}

Evidently, $75\%$ of instances having {\em `negligible'} hallucinations in \model\ and $56.3\%$ cases outperformed the most competitive model, MentalLlama. Consequently, $93.8\%$ of cases were deemed either {\em `totally relevant'} or {\em `relevant to some extent'} for practical applications by experts. However, with $12.25\%$ of instances marked {\em hallucinated}, there remains room for improvement in the future.

\paragraph{Impact.}\textcolor{black}{
The end goal of this study is to prepare \model\ for a pilot study for experts via an assistive framework that can help professionals streamline note-taking. Subsequently, this will allow more focus on patient care and improve time efficiency.}

\begin{table}[!t]
\centering
\resizebox{\columnwidth}{!}{
\begin{tabular}{c|cccccc}\toprule
& Af. Att. & Burden & Ethic. & Intv. Coh. & Cost & Percv. Eff. \\
\cmidrule{1-7}
$\mu$  & 1.23 & 0.36 & 1.03 &	1.20 & 0.85 & 1.27 \\
($\Delta$) & (0.21) & (0.34) & (0.23) & (0.33) & (0.19) & (0.27) \\
\bottomrule
\end{tabular}
}
\caption{Expert evaluation across six domain-centric metrics: {\em affective attitude (Af. Att.), burden, ethicality (Ethic), intervention coherence (Intv. Coh.), opportunity cost, and perceived effectiveness (Percv. Eff.)}. We present mean ($\mu$) and standard deviation ($\Delta$) scores.}
\label{tab:experteval}
\vspace{-5mm}
\end{table}

\section{Conclusion}
In our work, we explored the research space of mental health counseling summarization. Our research focused on strategically orchestrating LLMs to plan before generating content, thereby incorporating relevant dimensions of conversational structure- and domain-centric knowledge into the summarization process. We proposed a novel {\em planning engine} for LLMs, integrated and presented in this research with MentalLlama as \model. We compared the performance of \model\ against a spectrum of 14 potential baseline methods, including state-of-the-art LLMs like Mistral, Llama-2,  Flan-T5, Zephyr, and MentalLlama. Next, we presented an extensive evaluation spanning automatic, human, and expert evaluations. Expert analysis indicated that the summaries generated by \model\  exhibit better relevance and structure, often surpassing the closest competitive LLMs and, in certain instances, even surpassing the gold standard. We conclude by discussing our research's ethical considerations and applicability, emphasizing our approach as an assistive module tailored exclusively for mental health experts, which mitigates potential risks associated with direct client interactions, ensuring a safeguard against unintended harm. This strategic positioning underscores our commitment to safe utilization in mental health contexts.

% \clearpage

\section{Limitations}
Research in the space of mental health counseling summarization, in general, poses several complex challenges. Firstly, there's a substantial scarcity of diverse and high-quality datasets tailored for counseling summarization. Our research predominantly relies on the MEMO dataset. To the best of our knowledge, it is the only publicly available dataset. The limited dataset diversity hinders robustness, emphasizing the critical need for more datasets in this research area. Another challenge arises from the scale of the LLMs utilized in our research. Pretraining these super-large models on domain-specific corpora incurs significant costs and environmental implications. MentalLlama, chosen as the foundation model due to its pretraining corpus, underscores the necessity for more domain-centric models specifically tailored for mental health counseling. Additionally, evaluating LLM generations in a sensitive domain like mental health necessitates a nuanced understanding of the domain's intricacies. LLMs, by nature, exhibit tendencies to generate information that might not accurately represent the domain context and could potentially "hallucinate". Hence, our research, on one end, reduces hallucination by planning the generation and, on the other end, is positioned as an assistive tool tailored for mental health experts rather than direct client-facing applications. This approach mitigates the risks associated with LLM-generated content and underscores the need for expert oversight and intervention in sensitive mental health contexts.

\section{Ethical Considerations and Future Work}
Our research is an augmentation of established state-of-the-art foundation models, focusing on enhancing the generational proficiency of Language Models (LLMs) via planning LLM generation rather than introducing LLMs from scratch. This approach ensures continuous refinement, specifically targeting improvements in mental health counseling summarization within existing models. {\bf Given the sensitive nature of research in the domain of mental health, we portray our research as an assistive module designed exclusively for mental health experts. Such a strategic approach mitigates risks associated with direct support-seeking client interactions, safeguarding against unintended impacts on client sentiments or mental states}. Mental health professionals retain autonomy, choosing whether to accept \model's generated summary or further tailor it based on their expertise. However, our research scope acknowledges the evolving nature of complexities and diversity, emphasizing the need for not merely larger but planned LLMs. This leads to aligning LLMs with optimal parameter sizes and prioritizing the planned generational capabilities. Hence, paving the way for more responsible LLMs, upholding ethical standards in the research for sensitive domains like mental health counseling.

% \section 

\bibliography{custom}

\clearpage
\appendix

\section{Discussion on Structural vs Domain Knowledge} \label{strVSdom}
In this study, we underscore two fundamental sources of knowledge: understanding conversational structure and integrating domain-specific knowledge. This arrangement helps cover the entire context precisely and, at the same time, adds domain knowledge to the LLM generations. In our proposed method, \model, we incorporate structural information through sheaf learning and infuse domain knowledge through knowledge filtering-cum-scaffolding. 

\paragraph{Sheaf Learners.} To capture the entire structure of the dialogue, we initiate the construction of a dialogue graph. This initial step required the conversion of dialogue interactions into a graph representation, facilitating the subsequent embedding process for LLM. Graph representations are natively considered to be generated via Graph Neural Networks (GNNs); however, their performance diminishes notably in heterophilic settings (similar neighbor nodes). We apply a vector space structure of conversational graphs called sheaves to introduce the notion of geometry in the structural understanding of conversation. Sheaf learners, as discussed in Section \ref{method:structure}, hold structural context by learning each sheaf in an end-to-end fashion by learning the associated vector space with sheaf to understand geometry ({\em aka} conversational structure).

Yet, an intriguing question emerges -- {\bf \em Can sheaf learners also encapsulate domain knowledge?} The answer pivots on the breadth of knowledge required. Unlike conventional scenarios, our research explores both knowledge filtering and the integration of external domain-specific knowledge, necessitating the introduction of scaffolding.

\paragraph{Knowledge Scaffolding.} We modified the classical scaffolding method introduced by \citet{scaffold} and refined this method to align with conversational understanding intricately.  Our objective revolves around capturing filtered nuances and structuring them to provide domain-rich input to our LLM planner. This process ensures knowledge-rich delivery of information to the planner rather than fragmented knowledge post-filtering as done by \citet{10.1145/3534678.3539187}. This adaptation allows for the strategic incorporation of domain-specific knowledge, enhancing the LLM's planning process and enabling a more holistic understanding of the conversational context.

\begin{table}[t]
\centering
\resizebox{\columnwidth}{!}{
\begin{tabular}{lrrrr}\toprule
{\bf Model} &{\bf R-1} &{\bf R-2} &{\bf R-L} &{\bf BS} \\
\cmidrule{1-5}
Llama & 46.22 & 27.32 & 32.85 & 41.70 \\
MentalLlama & 48.90 & 28.16 & 30.44 & 41.00 \\
\rowcolor{blue!14} {\bf \model}\ ($fm:$ MentalLlama) & 48.05 & 28.91 & 33.27 & 42.76\\
\cmidrule{1-5}
$\Delta_{\model-BEST}(\%)$ & \textcolor{red}{$\downarrow 1.73$} & \textcolor{blue}{$\uparrow 2.59$} &\textcolor{blue}{$\uparrow 1.27 $} &\textcolor{blue}{$\uparrow 2.54$}\\
\bottomrule
\end{tabular}
}
\caption{\textcolor{black}{Results obtained on the ACI-BENCH clinical note generation dataset to assess the generalizability of \model. We report  Rouge-1 (R-1), Rouge-2 (R-2), Rouge-L (R-L), and Bleurt Score (BS).}}
\label{performanceComparison}
 \vspace{-4mm}
\end{table}

\begin{table*}
\centering
% \fontsize{6}{7}\selectfont
\resizebox{\textwidth}{!}{%
\begin{tabular}{l|p{30em}|p{30em}} \toprule
& \textbf{Gold Summaries} & {\centering \textbf{\model\ Summaries}} \\\cmidrule{1-3}
1 & {The patient is fed up with the job and is feeling anxious and fearful about taking responsibility. The patient's kids are doing drugs. The patient is conscious about kids. The patient prepared a report. Therapist provides a routine of CBT after reading report. The routine includes evaluating evidence of facts and carefully reviewing the facts, then resolving it. The patient finds no time for family and is anxious about kids. Therapist understood the negative thoughts of the patient. The therapist made the patient understand making a report is not going to work on kids, so they suggested making a good decision for welfare.}
&
{The patient expresses overwhelming self-doubt and anxiety in a recent professional situation. They question their ability to handle responsibility and fear the potential negative outcomes. Despite concerns, they also express reluctance to change jobs. The counselor employs CBT techniques to help the patient, highlighting the positive impact of their intervention. However, the patient continues to struggle with guilt. Ultimately, their anxiety stems from the power and responsibility of job. The counselor works to address these concerns, emphasizing professional responsibility.} \\
\hline
2 & {The patient almost quit smoking. The patient got late for a gig because he forgot to look at his to-do list. The therapist told patient to stick it on his door, so whenever the patient opens the door, he can look at his to-do list. The therapist reminded patient about the application for school. The patient started writing on a calendar to remember.}
&
{The patient reports significant progress in quitting smoking, noting three weeks without a cigarette and expressing improvements in taste and breathing. However, they mention a setback in their job search, having visited a restaurant during dinnertime for a potential gig, which they acknowledge as a short-sighted decision. The counselor offers practical advice, suggesting strategies to improve organization and memory, such as using reminders and checklists. They also discuss progress in the patient's research for schooling and encourage them to set application deadlines. The patient expresses gratitude for the support and suggests celebrating their success in quitting smoking with a treat for the counselor as well. Overall, the session focuses on acknowledging achievements, addressing setbacks, and providing tools for success in various aspects of the patient's life.} \\

\bottomrule
\end{tabular}
}

\caption{\textcolor{black}{Error Analysis: Comparison of the gold summaries with the \model\ summaries, highlighting key shortcomings. The table highlights two specific cases where the model tends to include exaggerated information in shorter dialogues and struggles to capture specific details and the patient's intrinsic traits, such as emotions and sentiments.}}

\label{tab:errorAnalysis}
\vspace{-5mm}
\end{table*}

We utilized the pre-set split ratio of 70:20:10 for the train, validation, and test set.

\section{Generalizability}
\textcolor{black}{It is worth noting that to the best of our knowledge, MEMO is the only publicly available dataset that directly fits into the definition of counseling summarization. However, in order to assess the generalizability of \model, we further add experiments on another clinical note generation dataset - the ACI-BENCH dataset. We present the results in Table \ref{performanceComparison}, evaluated across the same set of metrics (R1/2/L + BS) as presented in Table \ref{tab:results}. Our findings show that the proposed model surpasses the Llama model on the R1 metric by +3.95\% but shows a minute decline (0.85 R1 pts) when compared with MentalLama. On the other hand, our model surpasses Llama by +5.81\%, +1.27\%, and +2.54\% on R2, RL, and BS metrics, respectively. At the same time, our model surpasses MentalLama by +2.59\%, +9.29\%, and +4.29\% on R2, RL, and BS metrics, respectively.}

\section{Error Analysis}
\label{errorSection}
\textcolor{black}{We present our findings from a meticulous assessment of a sample of outputs using our proposed model, \model.}

\textcolor{black}{\paragraph{Finding 1:}  While the patient's overwhelming self-doubt and anxiety are mentioned, the specific details about their job dissatisfaction, concerns about their children, and intervention with CBT techniques are not adequately captured. Evidently, our model tends to include exaggerated information from the domain in shorter dialogues. In the case of Case 1, where only 20 utterances were present, the model included additional information beyond what was necessary. This highlights a tradeoff between incorporating external information and maintaining the relevance of generated summaries, marking it as a potential area for future study.}

\textcolor{black}{\paragraph{Finding 2:} In another case (Case 2), we observed the disparity in length between the gold and PIECE summaries. This is generally not an issue but we further observe that PIECE summarizes the dialogue with a length that doubles the brevity of the former (gold) and still conveys the same essential information. Furthermore, while appropriateness is generally maintained across both gold and PIECE summaries, there persists a notable deficiency in encapsulating the intrinsic traits of the patient, including emotions, tone, and sentiments, which are typically discerned and incorporated by therapists.}

\textcolor{black}{The error analysis highlights the conditional shortcomings in the PIECE summarization framework. We touched upon a few of these topics in Section 5.6 and Section 5.7; however, we understand the importance of explicitly including it in the paper. Hence, we commit to include a separate section on ‘error analysis’ in the camera-ready.}

\section{MEMO Dataset}
\label{memodef}
Drawing from the MEMO dataset source, we present the label definitions below:

\begin{itemize}
    \item Symptom and History (SH): Captures utterances that provide the most insightful information for the therapist to assess the patient’s situation.

    \item Patient Discovery (PD): In counseling sessions, patients often present with complex thoughts. The therapist builds a therapeutic relationship to help patients unravel these thoughts.

    \item Reflecting (RT): Therapist utterances are often concise to allow patients ample space to express themselves. Therapists may use imaginary scenarios to help understand the patient's perspective.

    \item Discussion Filler (DF): These are peripheral utterances in the conversation, such as pleasantries (‘Good morning!’), non-lexical fillers (‘Ummm’), acknowledgments (‘Right’), and restatements of affirmations (‘Yeah. Yeah’). These utterances carry little to no relevance in summary generation.
\end{itemize}

\section{Experimental Setup}

\subsection{Expert Evaluation Metrics}\label{expDef}
Experts followed an established framework as discussed in Section \ref{expVal}. The evaluation criteria involve the following metrics taken directly from the framework:

\begin{itemize}[leftmargin=*,noitemsep,nolistsep]
\item Affective Attitude: This dimension assesses how individuals feel about an intervention based on their clinical knowledge and perceptions. It examines subjective feelings toward the intervention.

\item Burden: The burden dimension measures the perceived effort required to participate in or understand the intervention. It considers factors such as spelling, grammar, and overall interpretation, reflecting the ease of use.

\item Ethicality: This dimension evaluates the extent to which an intervention aligns with an organization's value system and code of ethics. It explores whether there are any ethical concerns or conflicts with established ethical guidelines.

\item Intervention Coherence: Here, the focus is on understanding the intervention itself. It measures how well the intervention is comprehended, highlighting the clarity and coherence of the intervention's content or instructions.

\item Opportunity Cost: Opportunity cost assesses the benefits and drawbacks of using a particular intervention in a given context. It considers the potential gains and losses associated with adopting the intervention.

\item Perceived Effectiveness: This dimension evaluates how well the intervention is expected to perform in its intended setting. It assesses the perceived effectiveness of the intervention based on the expectations and experiences of individuals involved in its implementation.

\end{itemize}

\subsection{Quantitative Evaluation Metrics}
We employ ROUGE and Bluert scores. The computation of ROUGE scores is carried out using the \textit{py-rouge}  library\footnote{https://pypi.org/project/py-rouge/}, while Bleurt scores are calculated using the \textit{Hugging Face - Bleurt}  library\footnote{https://huggingface.co/spaces/evaluate-metric/bleurt}.

\subsection{Hardware and Parameters}
To accelerate model training and enhance performance, we utilized A100 GPUs, which significantly expedited the training process. The maximum load we deployed on GPUs was standard 7-billion parameter LLM models {\em viz.} Llama-7B, Mistral-7B, etc. We thoroughly explored the hyperparameter space to identify the optimal configuration for our model and present it in our supplementary code. \textcolor{black}{We used a full fine-tuning technique for our models. The learning rate was managed using a PyTorch scheduler, which gradually decayed the rate, starting from 1e-3 and decreasing by a factor of 0.1 with each epoch. The batch size for the training was set to 4, and the training was conducted over 10 epochs. Additionally, in all large language models (LLMs) used, only the last two layers were unfrozen during the fine-tuning process.}

\end{document}